\documentclass[fleqn,10pt]{wlscirep}
\usepackage[utf8]{inputenc}
\usepackage[T1]{fontenc}
\usepackage{bm}

\usepackage{comment}
\usepackage{amsmath}
\usepackage{makecell}
\usepackage{multirow}

\newcommand{\ignore}[1]{}

\title{High-entropy Advantage in Neural Networks' Generalizability}

\author[1, 2, *]{Entao Yang}
\author[3]{Xiaotian Zhang}
\author[3, 4]{Yue Shang}
\author[3, *]{Ge Zhang}
\affil[1]{Innovation Campus Delaware, Air Liquide, Newark, DE, USA}
\affil[2]{Department of Chemical and Biomolecular Engineering, University of Pennsylvania, Philadelphia, PA, USA}
\affil[3]{Department of Physics, City University of Hong Kong, Hong Kong, China}
\affil[4]{Department of Physics and Astronomy, University of Pennsylvania, Philadelphia, PA, USA}
\affil[*]{e-mail: entao.yang@airliquide.com; gzhang37@cityu.edu.hk}

\begin{abstract}

One of the central challenges in modern machine learning is understanding how neural networks generalize knowledge learned from training data to unseen test data. 
While numerous empirical techniques have been proposed to  improve generalization, a theoretical understanding of the mechanism of generalization remains elusive. 
Here we introduce the concept of Boltzmann entropy into neural networks by re-conceptualizing such networks as hypothetical molecular systems where weights and biases are atomic coordinates, and the loss function is the potential energy.
By employing molecular simulation algorithms, we compute entropy landscapes as functions of both training loss and test accuracy (or test loss), on networks with up to 1 million parameters, across four distinct machine learning tasks: arithmetic question, real-world tabular data, image recognition, and language modeling. 
Our results reveal the existence of \textit{high-entropy advantage}, wherein high-entropy network states generally outperform those reached via conventional training techniques like stochastic gradient descent.
This entropy advantage provides a thermodynamic explanation for neural network generalizability: the generalizable states occupy a larger part of the parameter space than its non-generalizable analog at low train loss.
Furthermore, we find this advantage more pronounced in narrower neural networks, indicating a need for different training optimizers tailored to different sizes of networks. 

\end{abstract}
\begin{document}

\flushbottom
\maketitle

\thispagestyle{empty}



\section*{Main}

\subsection*{Introduction}

Neural Networks (NNs) have shown remarkable success in solving a wide range of machine learning problems, from image recognition to natural-language processing. \cite{deng2009imagenet, vaswani2017attention, devlin2018bert, brown2020language} 
Despite their success, applications can face the problem of overfitting,\cite{srivastava2014dropout, Chiyuan2016, ying2019overview, zhang2021understanding} which occurs when an NN learns not only the underlying patterns in the training data, but also the noise and random fluctuations.
Consequently, the NN could perform exceptionally well on the training data but fail to generalize to new, unseen data.
The ability to generalize is a critical metric to evaluate an NN's usefulness in real-world applications, making 
\textit{generalizability} one of the central topics in the field. \cite{Chiyuan2016, zhang2021understanding} 
Modern NNs are usually overparameterized, i.e., the number of parameters exceeds the number of training data points.\cite{allen2019learning, zou2019improved}
Intuitively, an overparameterized model can easily overfit the training data, akin to how a high-order polynomial overfit a curve and fail to capture the underlying trend.
In fact, today's NNs have sufficient capacity to memorize the entire training set, as evidenced by their ability to easily fit data with random labels.\cite{Chiyuan2016}
However, despite their massive size, NNs have achieved state-of-the-art generalization performance in many different machine-learning tasks.\cite{he2016deep, silver2017mastering, jumper2021highly, merchant2023scaling}
Such surprising generalizability has sparked extensive research to understand its underlying mechanism. 

Analogous to the representation of all degrees of freedom in a physical system as a point in high-dimensional phase space, each specific  configuration of an NN, defined by the values of all its trainable parameters, can likewise be depicted as a point in a similarly high-dimensional parameter space. For an over-parameterized NN, there are many solutions in the parameter space that can fit the training data equally well.\cite{feng2023activity}
Some of the solutions generalize well, while others do not. 
Therefore, many studies on generalizability looked for its correlation with certain characteristics in the parameter space, including sharpness, the VC dimension, Rademacher complexity, weight norms, and intrinsic dimension.\cite{keskar2016large, jiang2019fantastic, fort2019goldilocks, ansuini2019intrinsic}
Unfortunately, this line of research has not yet generated convincing conclusions,\cite{jiang2019fantastic} as the best-performed measure, sharpness, has been challenged to be sufficient enough to determine the generalization by itself.\cite{dinh2017sharp, feng2023activity}
There are also studies attributing the generalizability to different regularization methods,\cite{srivastava2014dropout, neyshabur2014search, Chiyuan2016} which are protocols that bias the training trajectory toward some specific parts of the parameter space that have been empirically found to generalize well. 
Although empirical studies suggest that regularization can help generalization, many modern NNs can still generalize well without these methods, excluding them as the fundamental mechanism of generalization.\cite{Chiyuan2016} 

Here we investigate the mystery of NN generalizability from a physics perspective by conceptualizing it as a hypothetical molecular system. 
Each parameter in a NN (weights and biases) acts as an ``atom'' with its value representing its atomic coordinates.
We then compute the Boltzmann entropy of this  molecular system as a function of train loss and test performance (accuracy for classification or loss for regression).
Our results show that among all states that fits the training data well, the highest-entropy ones are highly generalizable. 
Specifically, we examined the entropy-generalizability relations in four distinct machine learning tasks, including arithmetic question, real-world tabular data, image recognition, and language modeling. 
Our results demonstrate that in all four tasks, the generalizability of high-entropy states are better or at least comparable with the states reached via the classical training optimizer like stochastic gradient descent (SGD). 
We refer to this as the \textit{high-entropy advantage} in this paper.
We intentionally  choose to compare with SGD training rather than other popular optimizers like Adam, because it has been shown empirically  that SGD can usually realize better generalization in neural networks training. \cite{keskar2017improving, zhou2020towards, gupta2021adam}
Recent work also shows that the noise of SGD introduces an additional effective loss (implicit regularization) which favors flat solution in neural networks' loss landscape and therefore can lead to better generalization.\cite{yang2023stochastic}
To better understand the condition of high-entropy advantage, we further studied its correlation with the network width. 
Our results indicate that wider NNs tend to have a smaller high-entropy advantage for the same task (e.g. reaching the limit of generalization), which agrees with derivations and the reported observations under the infinite network width limit.\cite{lee2018deep}

\subsection*{Results}
\subsubsection*{Arithmetic Question}
We begun our investigation from small fully-connected neural networks (FNNs) on a relatively simple arithmetic task: learning a predefined equation of binary classification.
Here we constructed a synthetic Spiral Classification Dataset, where every data entry is a point on 2-D space with a color (brown or green) determined by its coordinates.
We generated 20 green points and 20 brown points randomly, forming two spirals.
Our task is to predict a point's color given its coordination. Therefore, this is a binary classification problem with 2 input features (horizontal and vertical coordinates).
More details including the equation used and figure of this synthetic dataset are presented in the Supplementary Materials.

%
%

\begin{figure}[htb]
\centering
\includegraphics[width=\linewidth]{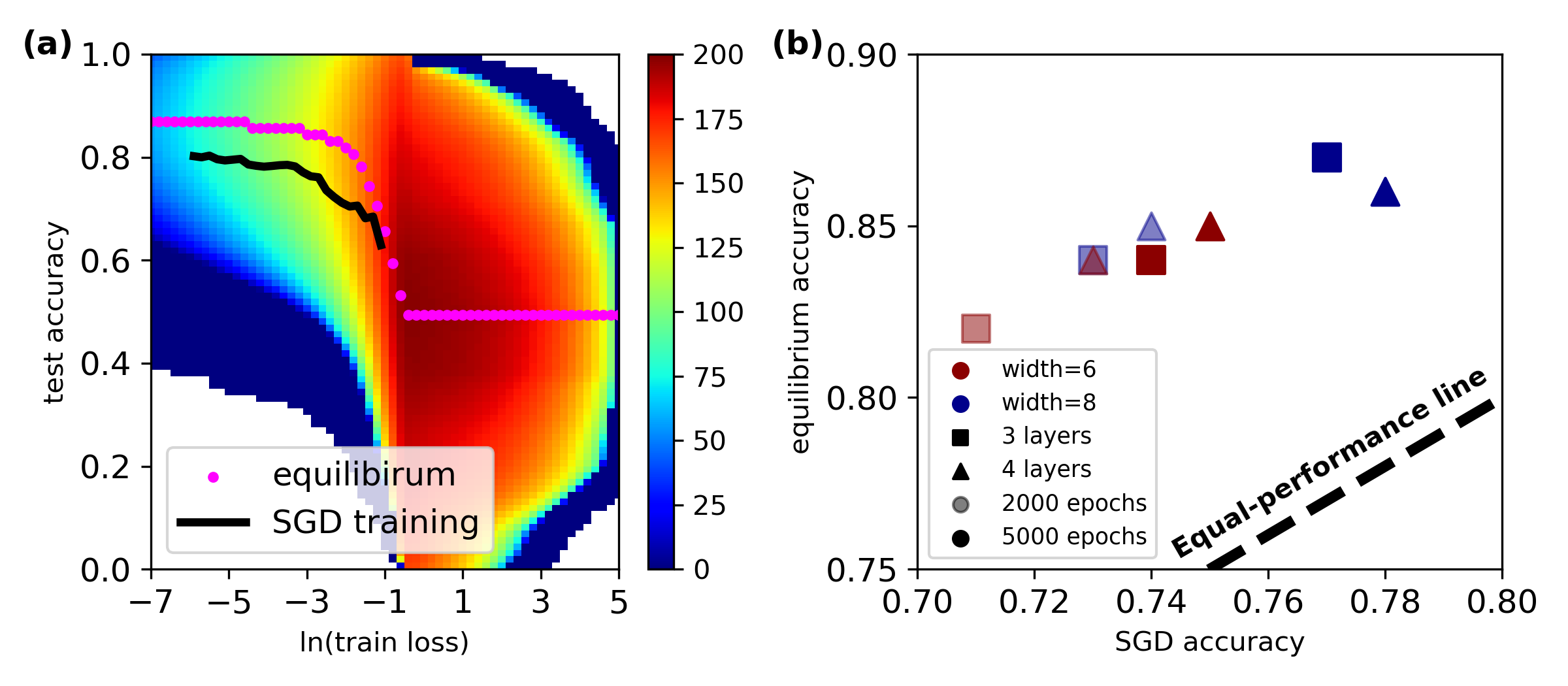}
\caption{Entropy analysis via WLMC for arithmetic question: (a) Entropy landscape as a function of $\ln(\mbox{train loss})$ and test accuracy. Color gradient represents the entropy gradient. (b) Equilbirum test accuracy versus the corresponding SGD test accuracy for $8$ different experiments.}
\label{fig:WLMC}
\end{figure}

We used Wang-Landau Monte Carlo (WLMC) method to sample the entropy landscape, $S(L_{train}, A_{test})$, as a function of the log-scaled train loss, $\ln{(L_{train})}$, and test accuracy, $A_{test}$.
The log-scaled train loss allows us to better sample the low $L_{train}$ states, corresponding to trained NNs.
In Fig.~\ref{fig:WLMC} (a), we present the entropy landscape for a ($3$ layers) $\times$ ($6$ neurons) FNN (116 trainable parameters).
Warmer color represents higher entropy.
The WLMC method (and the Wang-Landau Molecular Dynamics method we will employ later) calculates the entropy with an algorithm-dependent zero point; thus, the absolute value of the entropy lacks physical significance, but the entropy difference between any two points in Fig.~\ref{fig:WLMC} (a) carries physical meaning.
For each given $\ln(L_{train})$, we can calculate the corresponding equilibrium test accuracy:
\begin{equation}
\label{eq:euqi_accuracy}
    \langle A_{\mbox{test}}(L_{\mbox{train}}) \rangle=\frac{\int_0^1 a \exp\left[S(L_{\mbox{train}}, a)\right] da}{\int_0^1 \exp\left[S(L_{\mbox{train}}, a)\right] da}.
\end{equation}
where $S$ is the entropy and $a$ is variable standing for the test accuracy.
$\langle A_{\mbox{test}}(L_{\mbox{train}}) \rangle$ are plotted as magenta dots in Figure \ref{fig:WLMC} (a) and we refer it as the \textit{equilibrium accuracy} in this paper.

We then performed classical training via the SGD optimizer for 100 times and collected the $L_{train}$ versus $A_{test}$ trajectories. 
For each given $L_{train}$, we calculated the mean $A_{test}$, which is plotted as the black curve in Figure \ref{fig:WLMC} (a).
For simplicity, we refer it as the \textit{SGD accuracy} in this paper.
We can see that when $L_{train}$ reaches a low level [$\ln(L_{train}) \approx -0.5$)], the equilibrium accuracy increases rapidly with the decrease of $L_{train}$ and gets saturated when $\ln(L_{train}) \approx -3$.
In this regime, we observed a supremacy of the equilibrium accuracy over the SGD accuracy for each given $\ln(L_{train})$, which indicates the existence of high-entropy advantage. 
Another interesting observation is that when the training loss is at a high level ($\ln(L_{train}) > 0$), the equilibrium accuracy is around 50\%.
This also agrees with intuition, because when a FNN has a high $L_{train}$, the highest-entropy state of the model is just making random guesses, which results in 50\% accuracy for binary classification. 

We followed the same protocol and further tested on 3 additional FNN sizes and 2 different training time (total 8 experiments, see more details in Supplementary Materials).
Results are presented in Figure \ref{fig:WLMC} (b).
For all experiments, the training loss at the end of SGD training corresponds to an equilibrium test accuracy that outperforms the SGD accuracy with a large margin, further verifying that the high-entropy state can provide better generalization.
The smallest FNN tested here ($3$ layers $\times$ $6$ neurons) has 116 trainable parameters, which is still much larger than the number of training data points, 40, guaranteeing the overparamterized nature of all models.

\subsubsection*{Kaggle House Price Prediction}
After showing that high-entropy state can consistently provide better generalization for small FNNs, we then further extend the experiments to larger neural networks on real world tasks.
Due to the increase of parameters number in large networks, WLMC becomes inefficient, as it has a time complexity of $\mathbb{O}(n)$ to update one parameter, where $n$ is the total number of NNs' trainable parameters. 
Therefore, we employed the Wang-Landau Molecular Dynamics (WLMD) method, which updates all $n$ parameters together in each time step with time complexity $\mathbb{O}(n)$. (See Methods for more details)

We started from the House Price dataset\cite{de2011ames} on Kaggle website,\cite{house-prices-advanced-regression-techniques} where the task is to predict the the house price from its descriptors like number of bedrooms and house age.
For the sake of computational cost, here we only used the original training dataset which consists of 1460 houses.
Each house has 79 descriptors (excluding the house ID) and we performed one-hot encoding for all categorical descriptors, yielding a total of 331 descriptors / features for each data point eventually.
We randomly select 50\% data as the training set, with the remaining data reserved for testing.
A 2-layer FNN is used for this regression task, where the hidden layer has 20 neurons.
Therefore, the final model has 6661 trainable parameters, making it overparamterized.

\begin{figure}[htb]
\centering
\includegraphics[width=\linewidth]{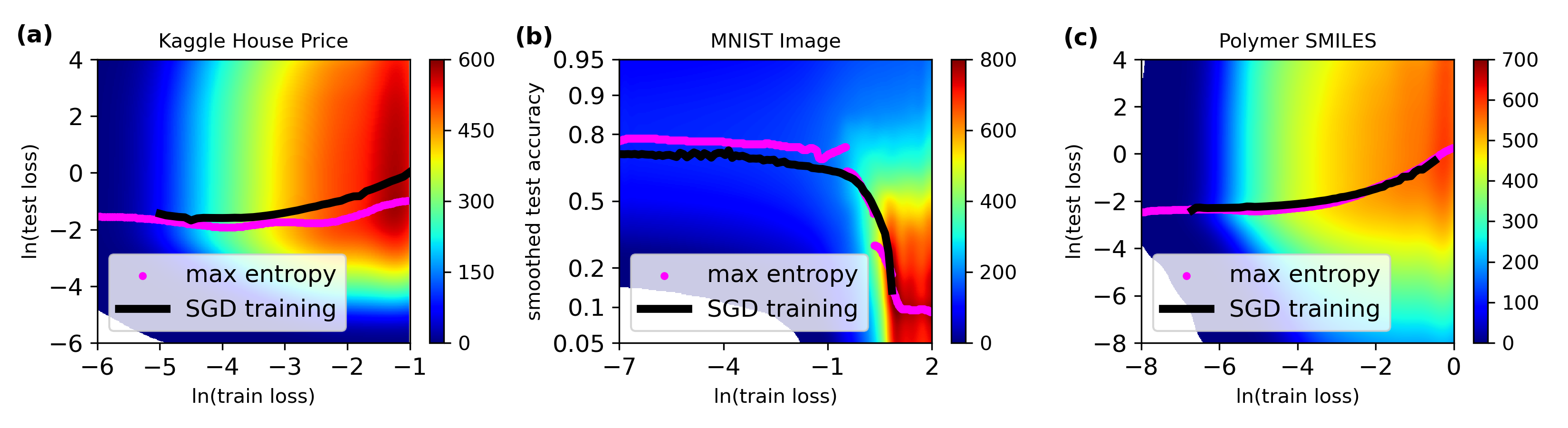}
\caption{Entropy landscape of (a) Kaggle Housing Price dataset; (b) selected MNIST dataset; (c) Polymer SMILES dataset. For the SGD training of all three models, the standard error at each train loss level is smaller than the width of the curve.}
\label{fig:WLMD}
\end{figure}

Following the similar protocol described above for the arithmetic question, we computed the entropy landscape, $S(L_{train}, L_{test})$,  as a function of train loss, $L_{train}$, and test loss, $L_{test}$.
Results are presented in Figure \ref{fig:WLMD} (a), where both $L_{train}$ and $L_{test}$ are on a log scale for clarity in the low-loss regime.
In statistical physics, the probability distribution over different states (different locations in the plot) is sharply peaked around the entropy maximum, making the max-entropy state thermodynamic equilibrium.
Therefore, we locate these max-entropy states at each $L_{train}$ (magenta dots), and compare with SGD trained states (black curve). 
SGD training results are averaged over 100 independent training instances using optimized hyperparameters, yielding error bars smaller than the black curve's width.
Our results suggest that at a given $L_{train}$, the corresponding max entropy loss is clearly lower than the test loss obtained via SGD training, demonstrating the high-entropy advantage in NNs generalizability.
More technical details, including data and hyperparameter tuning, are presented in Supplementary Materials.

\subsubsection*{MNIST Image: Handwritten Digit Recognition}
Modern neural networks have demonstrated remarkable success in computer vision.\cite{he2016deep}
One of the fundamental datasets used in the computer vision benchmark is the MNIST dataset, which consists of images with handwritten digits. \cite{lecun1998mnist}
Here we used it to further verify that the high-entropy advantage for the computer vision task.
Since many models have achieved 100\% accuracy on the test set of MNIST (see Leaderboard of Kaggle Digit Recognizer competition\cite{digit-recognizer}), we strategically increase the difficulty of the MNIST task by a 200 times reduction of the training dataset size.
This is done by randomly sampling 500 images from the original dataset and dividing them equally as the train and test sets, respectively.
We refer this dataset as the selected MNIST dataset in this paper.
The increased difficulty allows us to better observe the potential advantage of the high-entropy states.
We built a small convolutional neural network with 5 convolutional layers followed by one fully-connected layer.
This network has 362 trainable parameters, which is larger than the training dataset (250 images) in the selected MNIST, making it overparamterized.

In Figure \ref{fig:WLMD} (b), we presented the entropy landscape, $S(L_{train}, A^s_{test})$, of this neural network on the selected MNIST dataset.
Here $A^s_{test}$ is \textit{smoothed accuracy} which is differentiable, as required by the WLMD algorithm (See Methods for details).
We then find the max-entropy spot for each train loss on the landscape, which is again plotted as the magenta dots.
Similar to the landscape in Figure \ref{fig:WLMC} (a), we observed an increase of max entropy test accuracy when the train loss reaches a low level ($\ln(L_{train}) \approx-1$) and saturation at $\ln(L_{train}) \approx -4$.
When the train loss is high, we observed a high-entropy band around the test accuracy of 0.1.
This also meets our expectation because random guesses for this 10-class classification problem have an accuracy of 10\%.
We then performed classical training via the SGD optimizer for 200 times. (See more details in Supplementary Materials.)
The mean SGD accuracy are plotted as the black curve.
Our results suggest that for a given train loss, the corresponding max entropy accuracy generally outperforms the SGD training, especially when the train loss is small ($\ln(L_{train}) < -2$).

In order to confirm the existence of the entropy advantage in deeper neural networks, we investigated a 10-layer ResNet \cite{he2016deep} (43604 trainable parameters) on a selected CIFAR-10 dataset (5000 images, equally divided for training and testing). The results, presented in the Supplementary Materials, exhibit strong high-entropy advantage as well.

\subsubsection*{Polymer SMILES: Language Modeling}
Natural language modeling is another domain where neural networks have consistently outperformed traditional machine learning methods. \cite{vaswani2017attention, devlin2018bert, brown2020language}
Language modeling is also inherently differs from other machine learning tasks due to its reliance on semantic complexity and contextuality. \cite{liu2019linguistic}
Therefore, we further extended our tests to a language modeling task.
Specifically, we focused on a new emerging class of models called Chemical Language Models, which are language models trained on large databases of SMILES (Simplified Molecular Input Line Entry System) \cite{weininger1988smiles} strings and present promising performance in both predictive and generative tasks.\cite{flam2022language, born2023regression}
We utilized the TransPolymer model published recently, which achieves state-of-the-art performance in all ten different downstream tasks for polymer property prediction.\cite{xu2023transpolymer}
TransPolymer is a BERT (Bidirectional Encoder Representations from Transformers) family model\cite{devlin2018bert, liu2019roberta} and is pretrained on roughly 5 millions polymer SMILES augmented from the Pl1M database.\cite{ma2020pi1m}
This pretrained transformer-based language model is able to take the SMILES as input directly and generate a 768-dimension embedding for each given SMILES. 
The embedding is then fed into a regressor head, a fully-connected layer with SiLU activation function, and is regressed on different polymer properties.
Here we choose the Egb Dataset\cite{kuenneth2021polymer, xu2023transpolymer} to perform the entropy advantage experiment, which is the bandgap energy of bulk polymer and consists of 561 data points. 
This is also the dataset where TransPolymer has the best performance (test $R^2=0.93$), therefore, we can verify whether entropy advantage still exists for such a well-learned task.
80\% data is used for training and the remaining 20\% data is reserved for testing, which is same as the way reported in the original TransPolymer paper.\cite{xu2023transpolymer}

For sampling efficiency, we follow the fintuning strategy of large foundation model,\cite{chithrananda2020chemberta, ahmad2022chemberta, xu2023transpolymer} where the encoder embedding is fixed and only tuning the regressor head of the TransPolymer model. 
We also reduced the width of the regressor to 50 while keeping the SiLU activation function. 
Therefore, the final model has $38501$ parameters, making it overparamterized.
Entropy landscape of this language modeling task is presented in Figure \ref{fig:WLMD} (c).
The SGD training curve is obtained by performing the classical training for 40 times.
As a regression problem, we found that the corresponding max-entropy loss at each $L_{train}$ is slightly lower or comparable to its SGD training analog, which suggests that high-entropy state can generalize well.
In other words, even for a task that could be well-learned via the SGD training, there is still a high-entropy advantage.

\subsubsection*{Effect of Network Width}
Now that we have demonstrated high-entropy advantage in four distinct machine learning tasks, we will next study how this advantage is affected by the size of the neural network.
It is known that neural networks are equivalent to Gaussian processes in the limit of infinite width.\cite{neal1996priors, lee2018deep}
Therefore, it is reasonable to expect that the high-entropy advantage could also vary with the network width.

\begin{figure}[htb]
\centering
\includegraphics[width=0.7\linewidth]{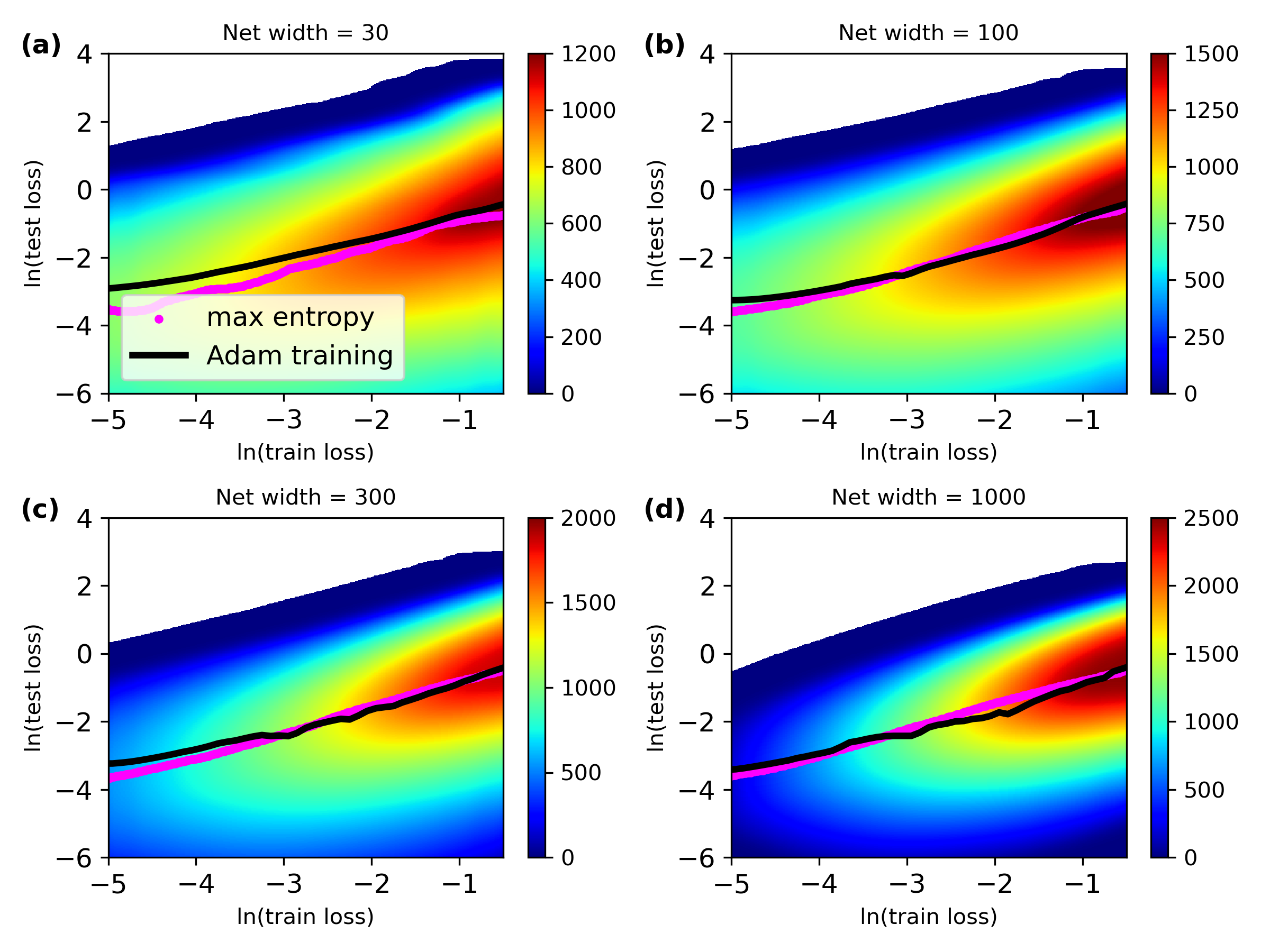}
\caption{Entropy landscape on spiral regression task with different NN width: (a) 30; (b) 100; (c) 300; (d) 1000.}
\label{fig:WLMD_width}
\end{figure}

To better evaluate this, we constructed a Spiral Regression Dataset (see Supplementary Materials for details) which consists of 500 data points and is divided equally for training and test sets.
Using the same WLMD method, we sampled the entropy landscapes on NNs with 2 hidden layers of four different widths, $W$, ranging from 30 to 1000. 
All four models are overparamerterized and the largest one has more than 1 million trainable parameters.
Our results, presented in Fig.~\ref{fig:WLMD_width}, suggest that the high-entropy advantage decreases as $W$ increases, and finally fades away when $W=1000$.
Note here we used the Adam optimizer here because the SGD optimizer performs much worse.
We also performed similar experiments on the Kaggle House price prediction, the selected MNIST image recognition, and the Polymer SMILES language modeling tasks.
These additional experiments confirmed a similar trend: the high-entropy advantage is more significant for narrower networks (see Figure \ref{fig:Sup_House} - \ref{fig:Sup_polymer} in the Supplementary Materials). 

\subsection*{Discussion}
In this paper, we investigated the relationship between the entropy and the generalizability of NNs. 
We preformed experiments in four distinctive machine learning tasks, covering arithmetic questions, real-world tabular data, computer vision, and language modeling.
WLMC and WLMD methods are used to sample the NNs' entropy landscapes for both classification [$S(L_{train}, A_{test})$] and regression tasks [$S(L_{train}, L_{test})$].
For classification tasks, at high train loss ($L_{train}$), the max-entropy states achieve random-guess accuracies, supporting the validity of the Wand-Landau-based algorithm in sampling the entropy of NNs; while at low train loss, the max-entropy states generally achieve higher accuracies than the SGD-trained states.
For regression tasks, the max-entropy states have lower or at least comparable test losses at all levels of train loss, comparing with their SGD training counterparts.
Therefore, for both categories of tasks, the max-entropy states exhibit promising generalizability, defining the high-entropy advantage in neural networks.

Considering entropy is the logarithm of the parameter-space volume, our findings demonstrate that the generalizable well-trained states occupy a \textit{larger} part of the parameter space than its non-generalizable analog, at the same train loss level. 
This surprising finding explains why training algorithms can find generalizable states, even in the absence of explicit regularization.
Furthermore, max-entropy states frequently surpass the generalization performance of SGD-derived states.
This suggests that while SGD (sometimes) favors parameters with marginally lower generalizability, the parameter space possesses a robust and extensive region of generalizable solutions, making the SGD-trained states frequently achieve excellent performance.

The contributions of our work are two-fold.
First, such a high-entropy advantage may be widely present in different NN architectures and learning tasks.
If so, our work shed lights on a new approach to enhancing model performance without changing its architecture: developing optimizers that can effectively find NNs' equilibrium (max entropy) states. 
This could be achieved by methods like low-temperature Newtonian molecular dynamics simulations or simulated annealing.
While a few studies have explored this direction before,\cite{engel1988teaching, el2021adaptive, surjanovic2022parallel} the existence of the high-entropy advantage in NNs underscores the great potential of further research.
Second, the high-entropy advantage also hints at a potential theoretical explanation for the generalizability of overparameterized NNs after SGD training.
The SGD training algorithm minimizes the loss function using an estimated gradient of the loss function, which presents strong anisotropy as demonstrated in recent work.\cite{yang2023stochastic}
The inaccuracy involved in such estimation may bias the trained NN toward not only flat but also higher-entropy minima, which possess higher generalizability.

Our results are also consistent with the philosophical principle of Occam's razor on a high level, which states that when competing theories of the same observation are presented, simpler ones should be preferred. 
For an NN with a fixed size, if it encodes a simpler theory of the training data, then fewer neurons are necessary for encoding, and more neurons can vary freely, contributing to a larger entropy. 
Our finding that these states generalize better provides a concrete reason why theoretical simplicity should be preferred; this also agrees with the observations that regularization can help generalization.

\ignore{
\textcolor{red}{Since now we use W=20 for house prediction task, we probably do not need this paragraph.}
We also noticed that the high-entropy advantage is less pronounced in the house price regression task and we speculate this may be due to NNs' lower efficacy with tabular-like data. \cite{grinsztajn2022tree, borisov2022deep}
In fact, benchmark studies have shown that tree-based model still achieves state-of-art performance for medium-sized tabular datasets ($\approx$ 10,000 data points). \cite{grinsztajn2022tree}
Recent studies attributed this to the irregular target functions resulting from the heterogeneous nature of tabular feature spaces, which biased the NNs to overly smooth solutions.\cite{Beyazit2023} 
Thus, NNs applied to these tasks are more likely to have a rugged entropy landscape, particularly in the low $L_{train}$ regime, making the sampling more challenging. 
Our WLMD method uses Gaussian smoothing in entropy updates and could be inaccurate in measuring rugged entropy landscapes. 
}

While there are many studies focusing on using machine learning models to solve problems in physics,\cite{raissi2019physics, udrescu2020ai, yang2022role, wang2023scientific, xiao2023identifying} the reverse directions retains much less attention. 
However, the foundation concepts of machine learning, especially for neural networks, are deeply rooted in physics. 
This historic connection is exemplified by the pioneering work including the Hopfield networks\cite{hopfield1982neural, hopfield1984neurons} and the Boltzmann machine,\cite{ackley1985learning, hinton1986learning} which is drawn directly from principles of statistical mechanics.
We believe our work (re)highlights the potential of treating NNs as physical systems, promoting  possibilities to apply well-established statistical physics methods to aid machine learning research.

\section*{Methods}
\subsection*{Wang-Landau Monte Carlo for Neural Networks}

Wand-Landau Monte Carlo (WLMC) is employed for the Spiral Classification Dataset and our results are averaged over 6 independently generated training and testing datasets.
For each dataset, 10,000 stages simulations has been performed, where each stage contains $3.2 \times 10^6$ steps. 
In each step, we randomly change one parameter in the network and then evaluate both train loss and test accuracy.
Therefore, WLMC algorithm has a time complexity of $\mathbb{O}(n)$ to update one trainable parameter, where $n$ is the total number of parameters in the network.
We also employed an adapted step size during the simulation by calculating the accepting rate for every 1000 steps.
The step size is increased by $10\%$ if the rate is higher than 0.7 and is decreased by $10\%$ when it is lower than 0.3.
The ``modification factor'' in original Wang-Landau method\cite{wang2001efficient} is set to be $f=\exp[5/(i+10)]$ for the i-th stage.

Parameters in the neural networks ranges from $-\infty$ to $\infty$, when no constraint is added.
Therefore, the volume of a parameter space is infinite, even for a tiny neural network. On the other hand, to make the thermodynamic quantities valid we must prescribe finite ranges for the parameters. 
According to early-established experience in neural network training, to alleviate the exploding-gradient and diminishing-gradient problems, the weight parameters should be on the order of $1/\sqrt{W}$.\cite{glorot2010understanding}
Thus, we limit parameters to the range of $[-2/\sqrt{W}, 2/\sqrt{W}]$ for our WLMC simulation.

\subsection*{Wang-Landau Molecular Dynamics for Neural Networks} 
When sampling the entropy landscape of practical neural networks with tens of thousands parameters, WLMC becomes too expensive as it needs to perform the entire forward propagation to update one parameter.
Therefore, we employed the Wand-Landau Molecular Dynamics (WLMD) method,\cite{kim2006statistical, junghans2014molecular} which has a time complexity of $\mathbb{O}(n)$ to update all parameters, where $n$ is the total number of trainable parameters in the network.
In addition, a Langevin thermostat is implemented to achieve the smooth temperature control during the simulation.
The parameters are constrained within the range of $[-2/\sqrt{W}, 2/\sqrt{W}]$ via reflecting boundary condition.
We also find that the range of the parameters does not qualitatively change our results within a reasonable range.

Unlike WLMC, the WLMD algorithm can only directly calculate the entropy as a function of differentiable variables. In order to calculate $S$ as a function of the test accuracy $A_{test}$, we designed a ``smoothed'' approximation of it:
\begin{equation}
    A^s_{test}=\mbox{Sig}(\alpha \times \delta l),
\end{equation}
where $\mbox{Sig}(x)=1/[1+\exp(-x)]$ is the sigmoid function, $\alpha$ is a hyperparameter, and $\delta l$ is the difference between the logit of the correct answer and the highest logit of incorrect answers. In the $\alpha \to \infty$ limit, we would have $A^s_{test}=A_{test}$. When $\alpha$ is finite, $A^s_{test}$ becomes a differentiable approximation of $A_{test}$. In practice we choose $\alpha=5$. This leads to a smoothing error, $|A^s_{test}-A_{test}|$, that is always smaller than $1.5\%$ except for the initial $10^7$ MD time steps. The initial part of the simulation have significantly higher smoothing errors (as much as $30\%$), but that does not affect the correctness of our final results since WLMD has the capability of starting from an incorrect entropy landscape and gradually converging to the correct entropy landscape.

The starting entropy $S(x, y)$, where $x=\ln(L_{{train}})$ and $y=\ln(L_{{test}})$ or $y=A^s_{test}$, is not zero but rather
\begin{equation}
    S_{{init}}(x, y)=c_i(x-x_{{max}})^2H(x-x_{{max}})+c_i(y-y_{{max}})^2H(y-y_{{max}}),
\end{equation}
where $c_i=3000$ is a constant and $H(x)=\begin{cases}
0, & x < 0\\
1, & x \ge 0
\end{cases}$ is the Heaviside step function. Parameters $x_{{max}}$ and $y_{{max}}$ are system-dependent and listed in the supplementary material. By starting with such an $S_{{init}}$, we make the algorithm concentrate its exploration effort in the $x<x_{{max}}$ and $y<y_{{max}}$ regime, improving efficiency. However, the entropy outside this regime becomes invalid, and are thus not presented in this paper.

For stable simulation, the scale factor in our simulation follows the schedule
\begin{equation}
    f(t)=\begin{cases}
\frac{t}{t_1}f_{\mbox{max}} & t\le t_1\\
f_{\mbox{max}} & t_1<t\le t_2\\
\frac{t_2-t_3}{t-t_3}f_{\mbox{max}} & t>t_2
\end{cases}
\mbox{,}
\label{eq:schedule}
\end{equation}
where $f_{\mbox{max}}$, $t_1$, $t_2$, and $t_3$ are parameters of the schedule. 
For each step, instead of adding one count to the corresponding position on the entropy landscape, a small Gaussian distribution is added to smooth the sampled entropy landscape.
Numerical details of all parameters employed for different machine learning tasks can be found in the Supplementary Materials.

We verified the correctness of our WLMD protocol by performing traditional (Newtonian) molecular dynamics simulations at a low temperature for the MNIST system. We observed that the test accuracy slowly increases and eventually reaches the high-entropy value found in WLMD simulations. This is consistent with the fact that a system must reach thermodynamic equilibrium in the infinite-time limit. More details are presented in the Supplementary Materials.


\bibliography{sample}


\section*{Acknowledgements}
The authors thank National Natural Science Foundation of China for supporting this research (Grant 12405043).
We also thank computational resources provided by Bridges-2 at Pittsburgh Supercomputing Center  through ACCESS allocation CIS230096.

\section*{Author contributions}
E. Y. and G. Z. conceived the research, performed simulations, and analyzed data. All authors participated in discussions, writing, and proofreading of the manuscript.

\section*{Competing interests}
The authors declare no competing interests. 

\newpage

\section*{Supplementary Materials}
\subsection*{Newtonian MD results for MNIST dataset}
To verify the correctness of the entropy diagram calculated from Wang-Landau MD simulations, we performed regular (Newtonian) MD of the MNIST system using the train loss as the potential energy. We observed that as the system evolves and approaches equilibrium, the test accuracy indeed increases and approaches the high-entropy value.

\begin{figure}[htb]
\centering
\includegraphics[width=0.48\linewidth]{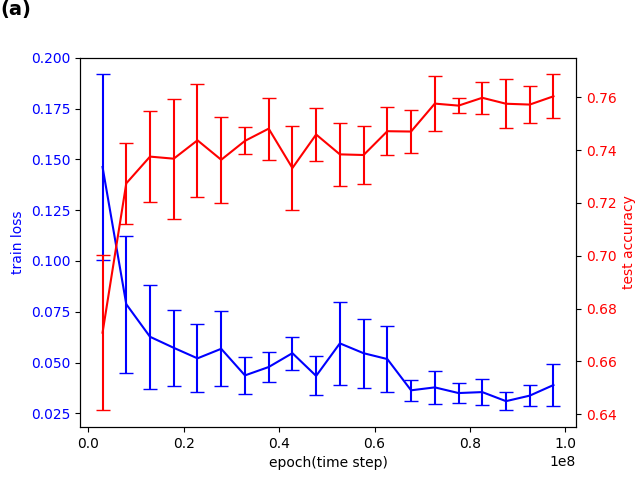}
\includegraphics[width=0.48\linewidth]{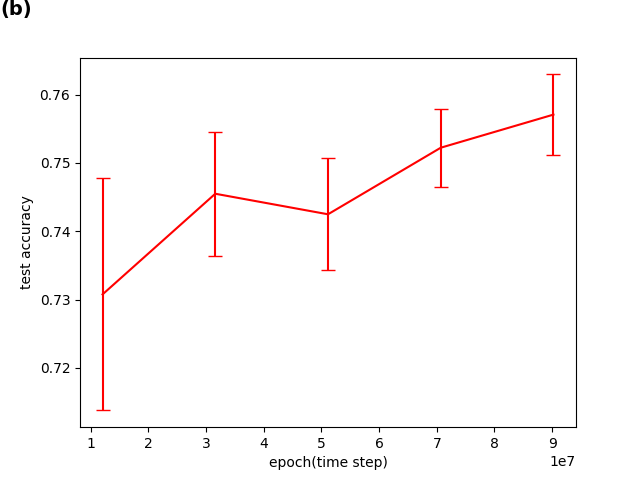}
\caption{(a) The train loss (energy) and test accuracy versus time steps (epoches) during Newtonian molecular dynamics simulations of the MNIST system. We averaged over 4 independent instances of simulations to reduce noise. (b) The test accuracy at time steps for which the train loss is in a small range of  $0.04<L_{\mbox{train}}< 0.041$.}
\label{fig:md}
\end{figure}

The MD simulation was performed at temperature $k_BT=0.005$ with Langevin thermostat, and the system eventually equilibrates at potential energy $E_p=L_{\mbox{train}}\approx 0.04$. From the entropy diagram (Fig.~2b), at this $L_{\mbox{train}}$ the equilibrium test accuracy is around 0.76, while SGD training achieved a test accuracy of 0.68. Our MD results are presented in Fig.~\ref{fig:md}. The test accuracy is indeed slowly increasing and approaching 0.76, but the train loss is also slowly decreasing. Since the test accuracy can increase due to either decreasing train loss or increasing entropy (as time passes), we further performed a separation-of-variables study by plotting the test accuracy for only the time steps with the train loss within a small window of $0.04<L_{\mbox{train}}< 0.041$. For this case, the test accuracy still increases over time, which is an evidence that the test accuracy is indeed positively correlated with entropy.

\newpage

\subsection*{Dataset Details}
\subsubsection*{Spiral Classification Dataset}

We generated the Spiral Classification dataset using the following equations:
\begin{equation}
\label{eq:sprial}
\begin{aligned} 
    &(x_{1i}, x_{2i})=(r_i \cos\theta_i+N_{1i}\mbox{, }r_i \sin\theta_i+N_{2i}) \\
    &\theta_i=2r_i+\pi y_i
\end{aligned}
\end{equation}
where $x_{1i}$ and $x_{2i}$ are the horizontal and vertical coordinates of a point, $r_i$ is uniformly distributed between 1 and 5, and $y_i$ is the label ({\it i.e.,} color) of a point.
To make it a binary classification task, the label is either 1 (brown) or 0 (green).
Both $N_{1i}$ and $N_{2i}$ are Gaussian random noises with mean 0 and standard deviation 0.1. 

\begin{figure}[htb]
\centering
\includegraphics[width=0.9\linewidth]{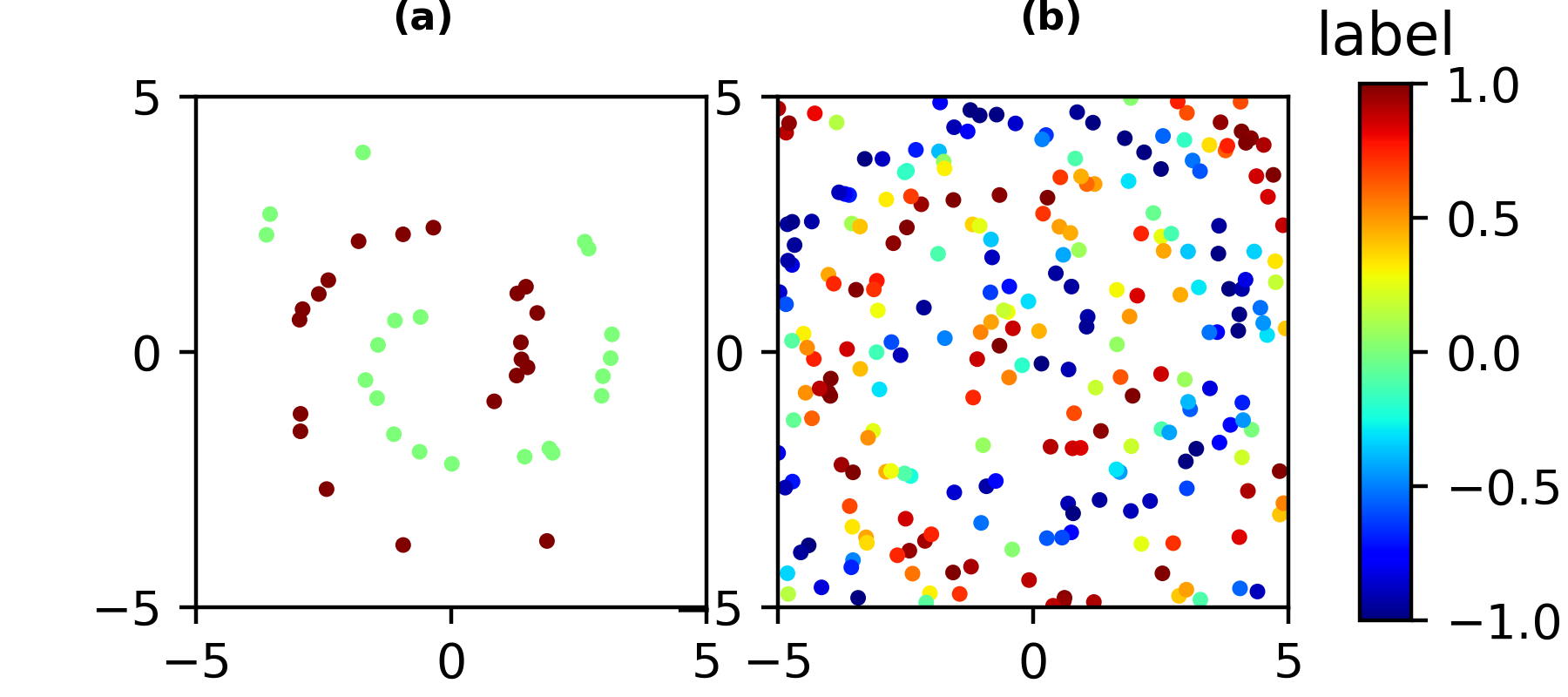}
\caption{(a) The Spiral Classification dataset. 
An independent test set was then generated following exactly the same procedure but with a different random seed. 
(b) The Spiral Regression dataset. Same as the Classification dataset, an independent test set with 250 points was generated.}
\label{fig:spiral}
\end{figure}

\subsubsection*{Spiral Regression Dataset}
We generated Spiral Regression datasets by first generating random coordinates, $-5<x_{1i}<5$ and $-5<x_{2i}<5$, for $N=50$, 100, or 250 points. We then calculate the label (color) of each point using the formula
\begin{equation}
y_i=\sin(\theta_i-2r_i),
\end{equation}
where $r_i$ and $\theta_i$ are the polar coordinates of the $i$th point.
The training datasets of both classification and regression tasks are presented in Figure \ref{fig:spiral}.

\subsubsection*{Kaggle House Price Dataset}
The Kaggle House Price dataset is collected from the Kaggle website.\cite{house-prices-advanced-regression-techniques}
Here we used the training data which consist of $1,460$ house sale prices with $79$ descriptors (like lot area and street type of the house). 
This tabular data is processed to better fit it into the neural network modeling.
All the descriptors are divided into two groups: numerical features and categorical features.
For numerical features, we performed standard scaling so that the data have a mean of 0 and a variance of 1.
For categorical features, we performed one-hot encoding.
After processing the raw data, we end up a new dataset which consists of $331$ features.
Therefore, our FNNs for the Kaggle House Price Prediction task have a input layer of 331 units.  
To make this learning task more difficult, we randomly selected $50\%$ data as the train set and the remaining $50\%$ data is used as the test set.

\subsubsection*{Selected MNIST Dataset}
The MNIST handwritten digits dataset is downloaded using the torchvision Python package. We randomly selected 250 images for the training dataset and 250 images for the test dataset. We greatly reduced the size of the dataset in order to increase the difficulty of machine learning and decrease the simulation time.

\subsubsection*{Polymer SMILES Dataset}
The Polymer SMILES dataset is collected from the TransPolymer work,\cite{xu2023transpolymer} which is the foundation chemical language model we used in this study.
The dataset is originally built by Ramprasad and coauthors, which consists of $561$ polymer SMILES strings and their bulk electric band gap energy (eV).\cite{patra2020multi}
TransPolymer takes SMILES strings directly into the encoder, which is pretrained over $5$ millions polymer SMILES string via mask language modeling.\cite{xu2023transpolymer}
This pretrained encoder can generate an embedding vector with a dimension of $768$ for each polymer SMILES, which is then fed into the regressor head for downstream task learning.
For sampling efficiency, we fixed the encoder of TransPolymer and only finetuning the regressor head, which is also a commonly employed strategy for language model fine-tuning.\cite{chithrananda2020chemberta, ahmad2022chemberta, xu2023transpolymer}
We follow the same protocol reported in the original TransPolymer work and randomly selected $80\%$ data for train set and the remaining $20\%$ data is used as test set.

\newpage

\subsection*{Neural network and training parameters}
To benchmark the max-entropy state, we trained the corresponding neural networks for each task via the SGD optimizer, except for the Sprial Regression task where SGD performs much worse than Adam. 
For a given network architecture, two hyperparameters have been tuned including learning rate and batch size.
Best hyperparamters for different neural networks can be found in Table \ref{tab:Hyper_Kaggle} to \ref{tab:Hyper_cifar}.
Note for Kaggle House Price Prediction task, three different batch sizes have been employed for all neural networks to collect enough data for the full range of train loss. 
To make sure we are comparing max-entropy state with the best-available SGD training state, we always select the lowest test loss value for SGD training if there are more than one network falls into the bin of a certain train loss level. 
After determining the best training hyperparameters, we trained $40\sim200$ neural networks and collected the train loss-test loss (regression task) or the train loss-test accuracy (classification task) trajectories during the entire training process of each task.
These trajectories are then used to calculate the SGD (Adam for Sprial Regression) training curves presented in Figure \ref{fig:WLMC} to \ref{fig:WLMD_width}.

\begin{table}[h!]
    \centering
    \begin{tabular}{|c|c|c|c|c|c|}
        \hline
        Network Width & \makecell{Number of \\ parameters} & Best LR & Best Batch Size & LR Optimized Range &  \makecell{Batch Size \\ Optimized Range} \\ \hline
        10 & 3331 & 5e-4   & 16  & [1e-4, 1e-2] & [16, 128] \\  \hline
        20 & 6661 & 5e-4   & 16 & [1e-4, 1e-2] & [16, 128]  \\ \hline
        50 & 16651 & 5e-3   & 16  & [1e-4, 1e-2] & [16, 128] \\ \hline
        100 & 33200 & 8e-3  & 64  & [1e-4, 1e-2] & [16, 128] \\ \hline
    \end{tabular}
    \caption{Hyperparameters tuning details for Kaggle House Price Prediction}
    \label{tab:Hyper_Kaggle}
\end{table}

\begin{table}[h!]
    \centering
    \begin{tabular}{|c|c|c|c|c|c|}
        \hline
        Network Width & \makecell{Number of \\ parameters} & Best LR & Best Batch Size & LR Optimized Range &  \makecell{Batch Size \\ Optimized Range} \\ \hline
        2 channels & 362 & 0.003   & 4  & [0.001, 0.1] & [4, 256] \\  \hline
        3 channels & 646 & 0.003   & 4  & [0.001, 0.1] & [4, 256] \\  \hline
        4 channels & 1002 & 0.003   & 4  & [0.001, 0.1] & [4, 256] \\  \hline
    \end{tabular}
    \caption{Hyperparameters tuning details for MNIST Digit Recognition}
    \label{tab:Hyper_MNIST}
\end{table}

\begin{table}[h!]
    \centering
    \begin{tabular}{|c|c|c|c|c|c|}
        \hline
        Network Width & \makecell{Number of \\ parameters} & Best LR & Best Batch Size & LR Optimized Range &  \makecell{Batch Size \\ Optimized Range} \\ \hline
        2 & 1541 & 5e-3   & 32  & [1e-4, 1e-1] & [16, 128] \\  \hline
        10 & 7701 & 1e-2   & 32 & [1e-4, 1e-1] & [16, 128]  \\ \hline
        50 & 38501 & 1e-2   & 64  & [1e-4, 1e-1] & [16, 128] \\ \hline
        100 & 77001 & 1e-2   & 64  & [1e-4, 1e-1] & [16, 128] \\ \hline
    \end{tabular}
    \caption{Hyperparameters tuning details for Language Modeling on Polymer SMILES}
    \label{tab:Hyper_SMILES}
\end{table}

\begin{table}[h!]
    \centering
    \begin{tabular}{|c|c|c|c|c|c|}
        \hline
        Network Width & \makecell{Number of \\ parameters} & Best LR & Best Batch Size & LR Optimized Range &  \makecell{Batch Size \\ Optimized Range} \\ \hline
        30 & 1051 & 2e-3   & 5  & [5e-5, 2e-2] & [5, 25] \\  \hline
        100 & 10501 & 5e-3   & 10  & [5e-5, 2e-2] & [5, 25] \\  \hline
        300 & 91501 & 2e-3   & 5  & [5e-5, 2e-2] & [5, 25] \\  \hline
        1000 & 1005001 & 2e-3   & 10  & [5e-5, 2e-2] & [5, 25] \\  \hline
    \end{tabular}
    \caption{Hyperparameters tuning details for spiral regression. For this dataset, we tuned the hyperparameters using the largest dataset ($N=250$) only since the smaller datasets yielded poorer results. We also switched to Adam optimizer since SGD yielded poor results.}
    \label{tab:Hyper_spiral}
\end{table}

\begin{table}[h!]
    \centering
    \begin{tabular}{|c|c|c|c|c|c|}
        \hline
        Network Width & \makecell{Number of \\ parameters} & Best LR & Best Batch Size & LR Optimized Range &  \makecell{Batch Size \\ Optimized Range} \\ \hline
        6-12 channels  & 43604  & 1e-4   & 64  & [1e-5, 2e-2] & [32, 256] \\  \hline
    \end{tabular}
    \caption{Hyperparameters tuning details for CIFAR dataset. For this dataset, we did not experiment with different neural-network widths due to high computational cost.}
    \label{tab:Hyper_cifar}
\end{table}

\newpage

\subsection*{WLMD Simulation Parameters Summary}

\begin{table}[h!]
    \centering
    \begin{tabular}{|c|c|c|c|c|c|}
        \hline
        WLMD Parameters &  Kaggle House & MNIST & Polymer SMILES & CIFAR &   \makecell{Spiral \\ Regression} \\ \hline
        
        Gaussian Sigma    & 0.5   & 0.2  & 0.5 & 0.3 & 0.2 \\  \hline
        
        Gaussian Cutoff    & 2   & 0.8 & 2 & 1.2 & 0.8 \\ \hline
        
        Max $\ln($Train Loss$)$    & -1.0 & 2.0  & 0.0 & 1.0 & -0.5 \\ \hline
        
        \makecell{Max $\ln($Test Loss$)$ \\ or smoothed Accuracy} & 5.0 & 0.95  & 5.0 & 1.0 & \makecell{$\infty$ \\ (no constraint)} \\ \hline

        $f_{\mbox{max}}$    & 20  & 3  & 20 & 20 & 5 \\ \hline
        
        $t_1$ & 5e5   & 2e5  & 1e6 & 3e5 & 2e5 \\ \hline
        
        $t_2$ &  5e5 & 1e7  & 1e6 & 2e6 & 1e7 \\ \hline
        
        $t_3$ & 0 & 8e6  & 0 & 0 & 8e6 \\ \hline

        Timestep & 1e-4   & 3e-6  & 3e-5 & 2e-5 & 5e-5 \\ \hline   

        \makecell{Langevin Thermostat \\ Friction Coefficient} & 0.01   & 1e-5  & 0.01 & 1e-4 & 3e-5 \\ \hline   
        
    \end{tabular}
    \caption{WLMD Simulation Parameters for Different Machine Learning Tasks.}
    \label{tab:Hyper_WLMD}
\end{table}

\subsection*{Entropy Advantage for ResNet on selected CIFAR dataset}

\begin{figure}[htb]
\centering
\includegraphics[width=0.5\linewidth]{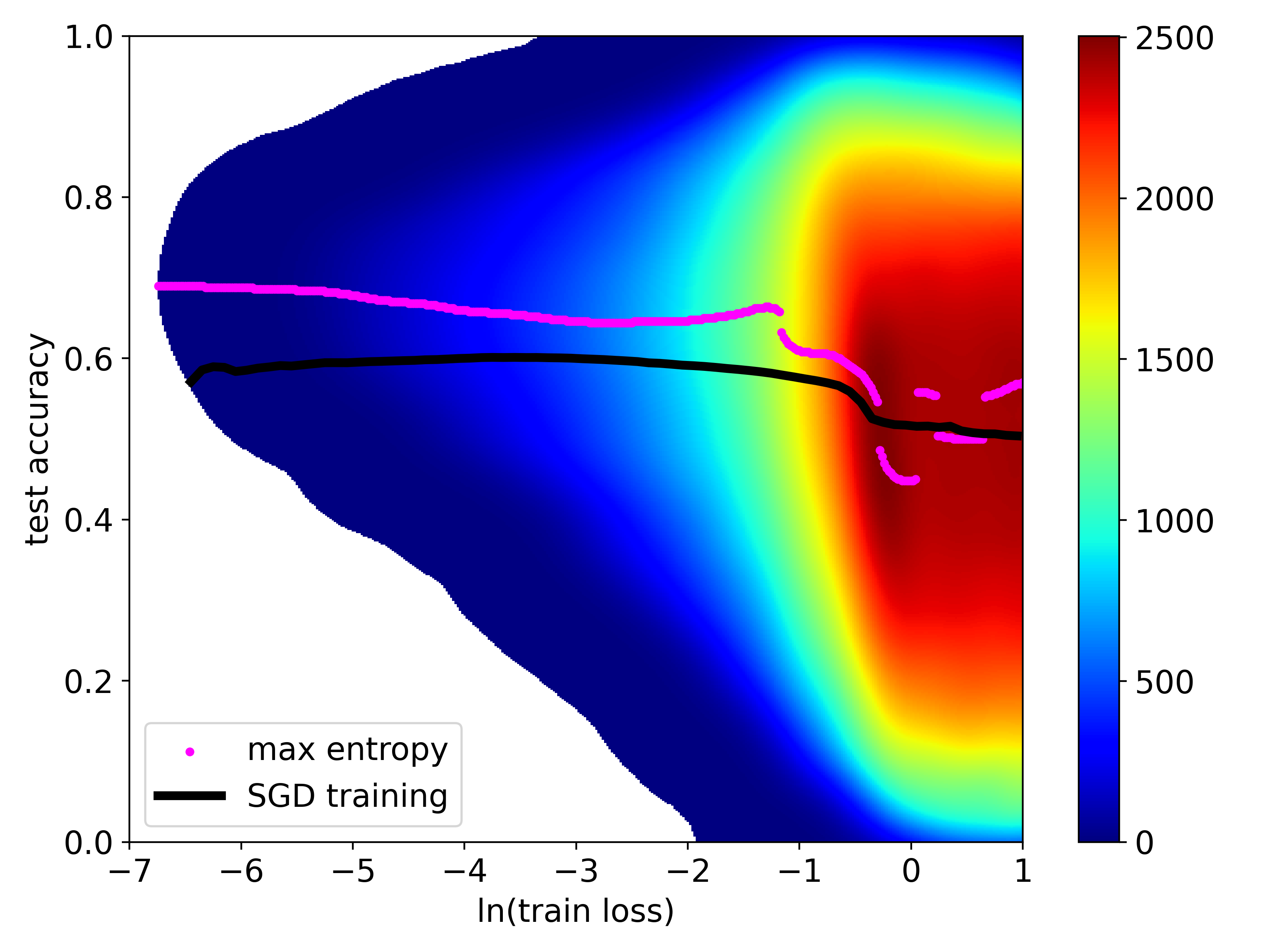}
\caption{Entropy landscape of ResNet on the selected CIFAR dataset}
\label{fig:Sup_CIFAR}
\end{figure}

To confirm the existence of entropy advantage in deeper neural networks with more complex architectures, we also constructed a 10-layer ResNet.\cite{he2016deep} The neural network consists of an initial convolutional layer with 6 channels, and then 4 residual blocks with 6, 12, 12, and 12 channels each, and a final fully-connected layer. The original ResNet architecture contains batch-normalization operations, which breaks the rigorous analogy between training losses and physical energy landscapes. Therefore, we replaced all of them with layer normalization operations. Our training and test datasets each contain 2500 randomly chosen images of either a deer or a frog from the CIFAR-10 dataset. Due to the significantly higher computational cost, we only ran WLMD for $3\times10^7$ time steps for this system. Nevertheless, our result in Fig.~\ref{fig:Sup_CIFAR} clearly shows an entropy advantage.

\newpage

\subsection*{High Entropy Advantages Analysis} 
\begin{figure}[htb]
\centering
\includegraphics[width=0.7\linewidth]{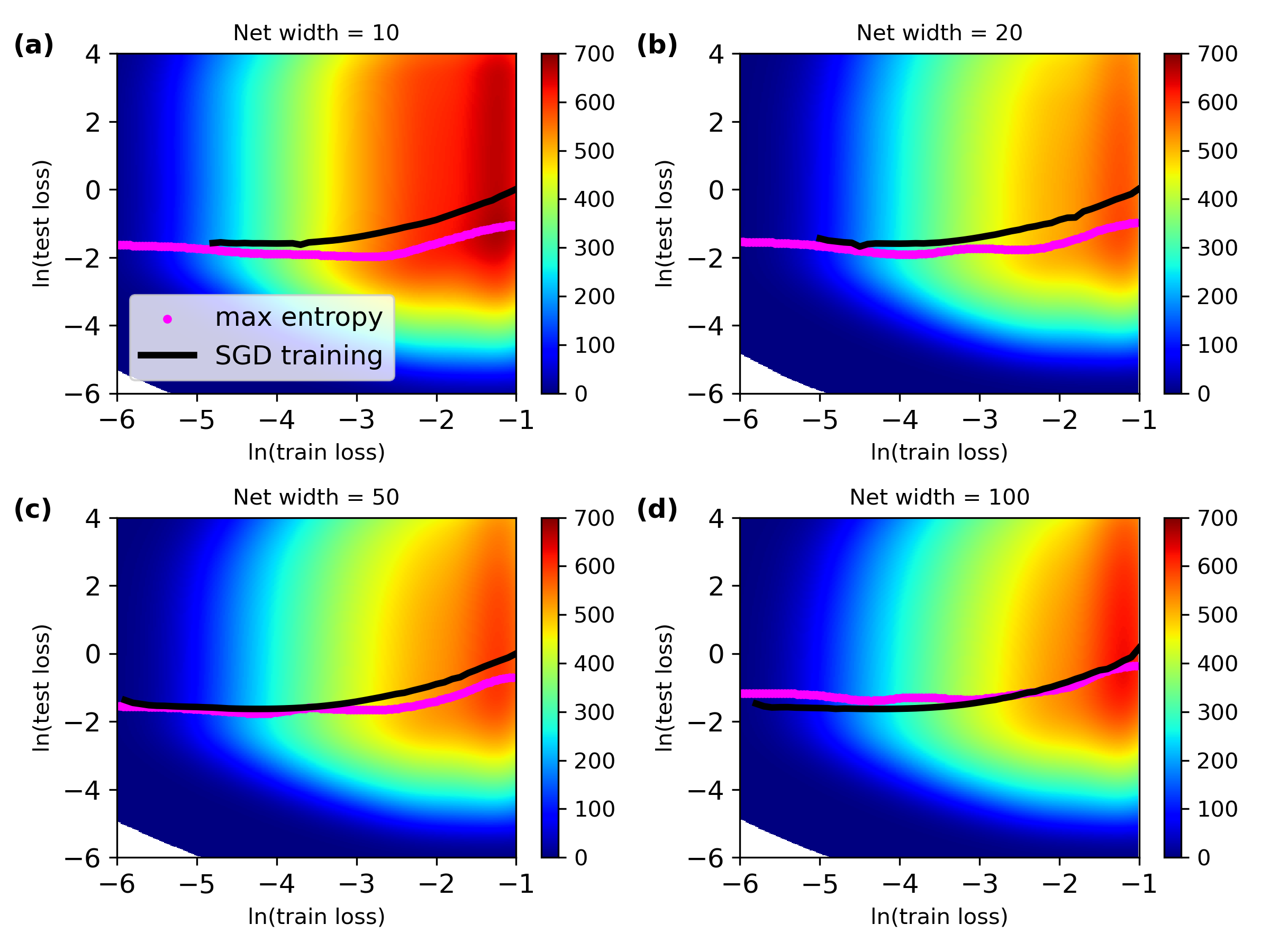}
\caption{Entropy landscape of Kaggle House Price prediction task with different neural network width, $W$: (a) $W=10$; (b) $W=20$; (c) $W=50$; (d) $W=100$.}
\label{fig:Sup_House}
\end{figure}

As presented in Figure \ref{fig:WLMD_width}, the high entropy advantage is more significant when the neural network is narrower, on the spiral regression task.
To further validate this observation, we performed same tests on the other three tasks including Kaggle House Price prediction, MNIST digit recognition, and Polymer SMILES Language modeling.
We found the similar trend in all three tasks: narrow networks present larger high-entropy advantage.
Detailed results are presented in Figure \ref{fig:Sup_House} to \ref{fig:Sup_polymer}. 

\subsubsection*{Kaggle House Price Prediction}

For Kaggle House Price prediction, we tested four different neural networks.
Each network has 1 hidden layer, where the width, $W$, varies from 10 to 100.
Our experiments suggest that the gap between the max-entropy curve and the SGD training curve becomes smaller (and almost disappeared at $W=100$) with the increase of $W$. 

\newpage

\subsubsection*{MNIST Digit Recognition}
\begin{figure}[htb]
\centering
\includegraphics[width=\linewidth]{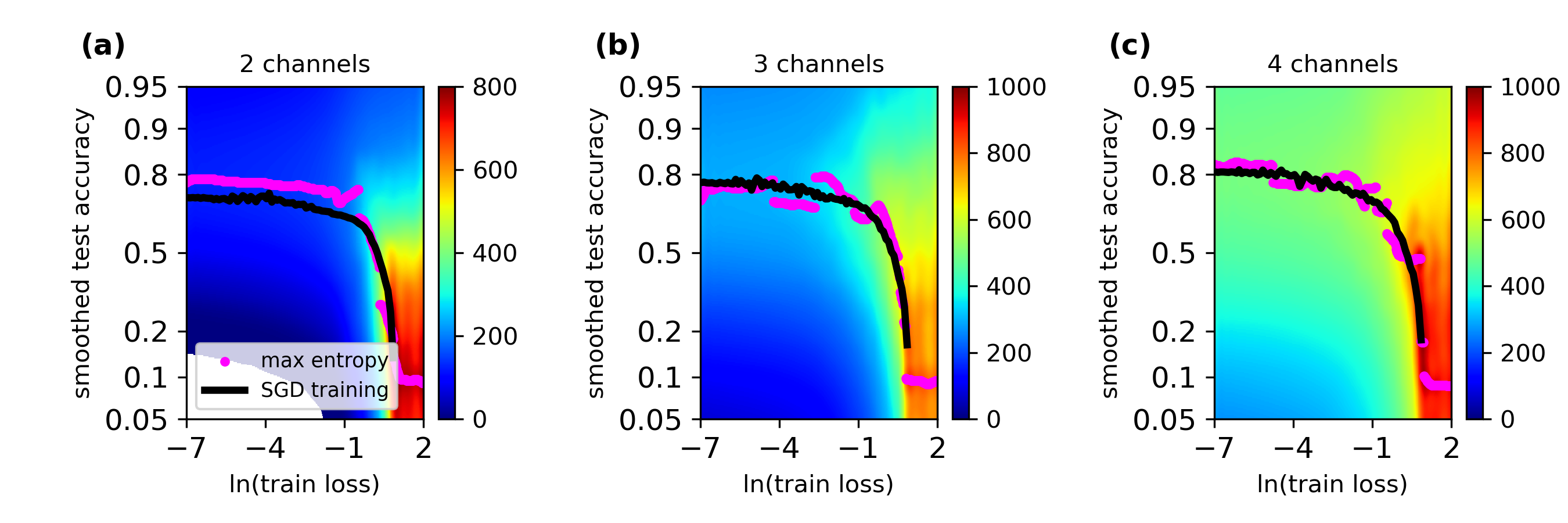}
\caption{Entropy landscape of MNIST Digit Recognition task with different channel numbers: (a) 2; (b) 3; (c) 4.}
\label{fig:Sup_MNIST}
\end{figure}

CNN is used in our experiment on the MNIST Digit Recognition task. 
Therefore, to evaluate the effect of network width on the high entropy advantage, we varied the channel number and tested three different CNNs.
Results are presented in Figure \ref{fig:Sup_MNIST}.
While CNNs with 3 and 4 channels present similar level of advantage, the max-entropy state of the 2-channel CNN has clearly higher test accuracy when train loss is small. 

\newpage

\subsubsection*{Polymer SMILES Language Modeling}
\begin{figure}[htb]
\centering
\includegraphics[width=0.7\linewidth]{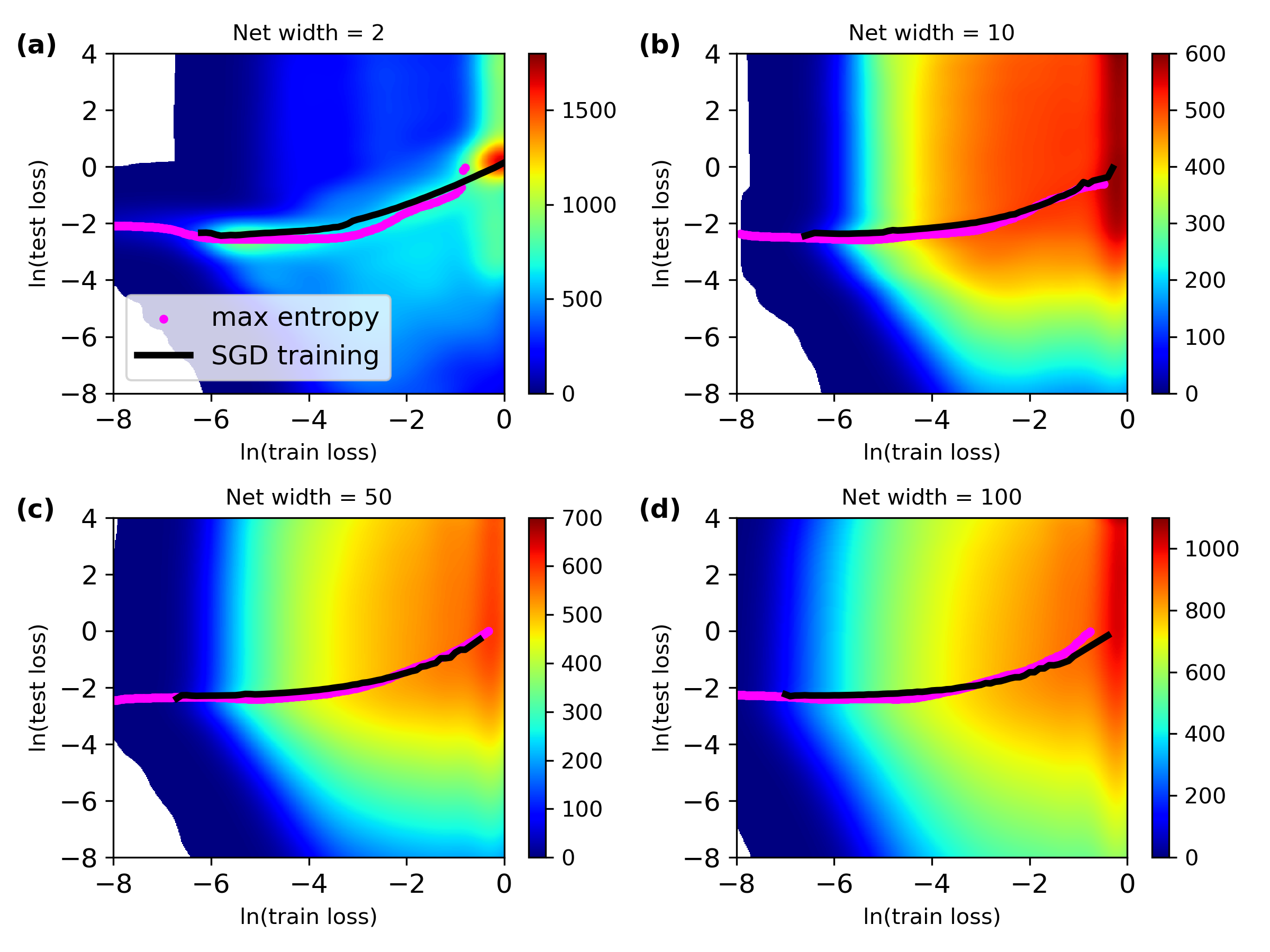}
\caption{Entropy landscape of Polymer SMILES language modeling with different regressor head widths, W: (a) W=2; (b) W=10; (c) W=50; (d) W=100. }
\label{fig:Sup_polymer}
\end{figure}

\begin{figure}[htb]
\centering
\includegraphics[width=0.7\linewidth]{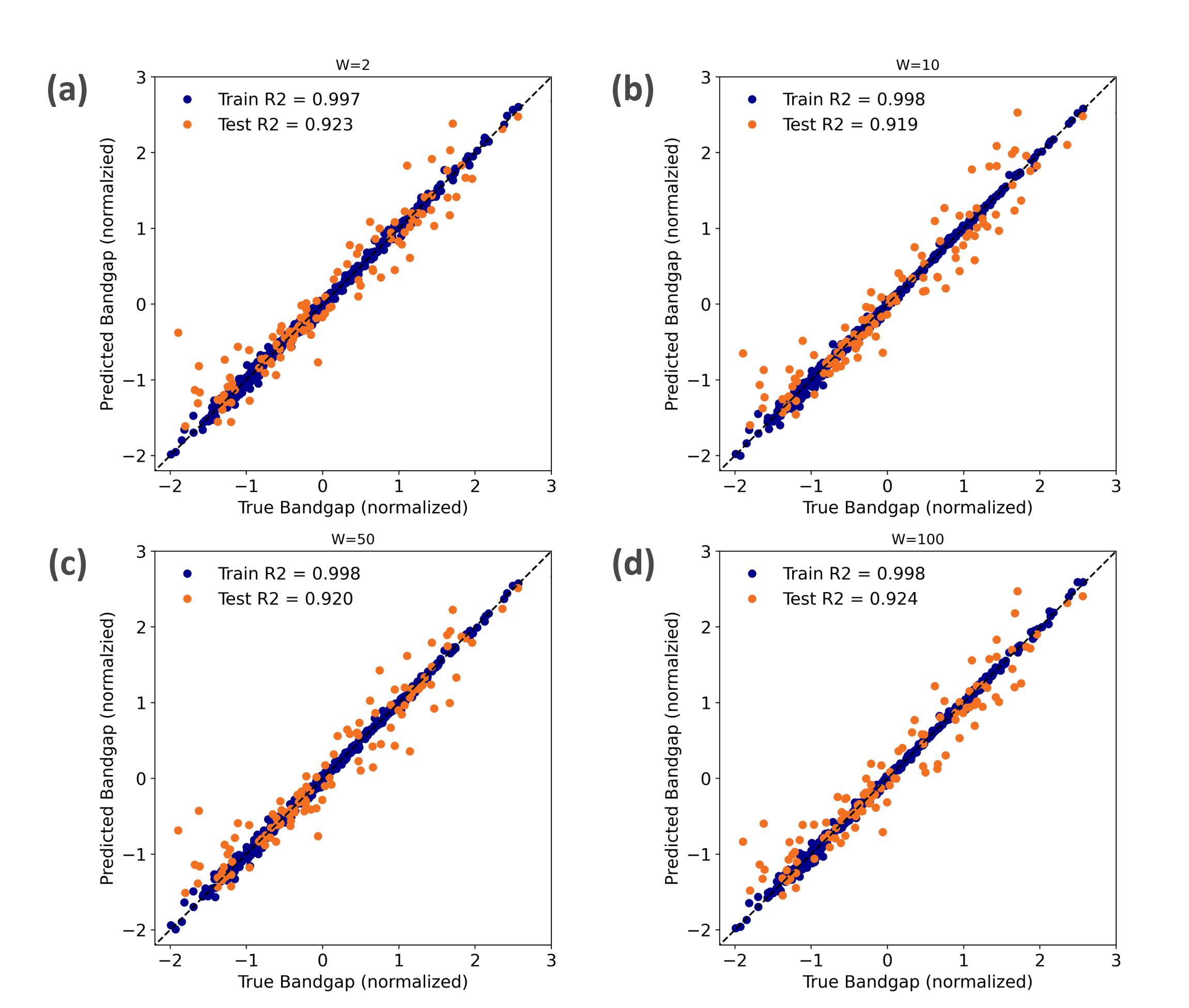}
\caption{SGD training results for language modeling with different regressor head widths, W: (a) W=2; (b) W=10; (c) W=50; (d) W=100. Test $R^2 > 0.9$ (orange points) for all four regressor head we tested, suggesting this is a regression task that can be well-learned by neural networks.}
\label{fig:Sup_polymer_sgd}
\end{figure}

We focused on the regressor head of the TransPolymer\cite{xu2023transpolymer} model in Polymer SMILES language modeling task, which is a linear layer.
Therefore, we varied its width from 2 to 100 to evaluate the high entropy advantage at different network width .
As we mentioned in the main text, this task is reported to be well-learned. \cite{xu2023transpolymer}
Our results further verify this as the test $R^2$ exceeded 0.9 for all four regressor heads we trained using SGD (see Figure \ref{fig:Sup_polymer_sgd}).
However, max-entropy state is still slightly better than the SGD training state when $W=10$ (Figure \ref{fig:Sup_polymer} (b)).
For increased width $W=50$ and $W=100$, the max-entropy state performs \textit{equally well} with SGD training in at all training losses (Figure \ref{fig:Sup_polymer} (c) and (d)).
When we use an extremely narrow network where $W=2$, we can see a larger gap between the max-entropy curve and the SGD training curve. (Figure \ref{fig:Sup_polymer} (a))
While the entropy landscape of in Figure \ref{fig:Sup_polymer} (a) looks different from the others, two independent experiments have been ran to make sure that it is not due to randomness.
We speculate this distinct shape of entropy landscape is due to the extreme narrow regressor head employed ($W=2$). 

\end{document}